\documentclass[journal]{IEEEtran}

\usepackage{amsmath,graphicx}
\usepackage{color}
\usepackage{graphicx}
\usepackage{subfig}

\newcommand{\vect}[1]{\boldsymbol{#1}}

\begin{document}
%
\title{MR to X-ray Projection Image Synthesis}
%
%
%

\author{Bernhard Stimpel, Christopher Syben, Tobias W\"urfl, Katrin Mentl, Arnd D\"orfler, and Andreas Maier%
\thanks{This work has been supported by the project P3-Stroke, an EIT Health innovation project. EIT Health is supported by EIT, a body of the European Union.}
\thanks{B. Stimpel, C. Syben, T. W\"urfl, K. Mentl, and A. Maier are with Friedrich-Alexander-Universit\"at Erlangen-N\"urnberg, Pattern Recognition Lab, Erlangen, Germany.}
\thanks{B. Stimpel, C.Syben, and A. D\"orfler are also with Friedrich-Alexander-Universit\"at Erlangen-N\"urnberg, Department of Neuroradiology, Erlangen, Germany.}

}

\maketitle
\pagestyle{empty}
\thispagestyle{empty}

\begin{abstract}
Hybrid imaging promises large potential in medical imaging applications. To fully utilize the possibilities of corresponding information from different modalities, the information must be transferable between the domains. In radiation therapy planning, existing methods make use of reconstructed 3D magnetic resonance imaging data to synthesize corresponding X-ray attenuation maps. In contrast, for fluoroscopic procedures only line integral data, i.e., 2D projection images, are present. The question arises which approaches could potentially be used for this MR to X-ray projection image-to-image translation. We examine three network architectures and two loss-functions regarding their suitability as generator networks for this task.
All generators proved to yield suitable results for this task. A cascaded refinement network paired with a perceptual-loss function achieved the best qualitative results in our evaluation. The perceptual-loss showed to be able to preserve most of the high-frequency details in the projection images and, thus, is recommended for the underlying task and similar problems.
The abstract goes here.
\end{abstract}


\section{Introduction}
\label{sec:intro}
Promising concepts on how a combined magnetic resonance (MR) and computed tomography (CT) imaging device may look like were proposed in the past. Wang et al. \cite{Wang2013} published a top-level design of an MR-CT scanner consisting of two superconducting electromagnets surrounding multiple, rotatable X-ray sources. The desired application for their model is combined image reconstruction for plaque characterization. In contrast, \cite{Fahrig2001} focused on the interventional applicability of a hybrid MR-X-ray system and showed the great potential of this application. 
Assuming an imaging device that is capable of acquiring corresponding X-ray and MR projection images simultaneously, or at least consecutively in the same state of motion,  the combined information would be highly useful for fluoroscopic procedures. On the one hand, overlay strategies of both modalities in their respective form could be used to simultaneously visualize soft- and dense-tissue or -material. On the other hand, the information of one modality could be transferred to the domain of its counterpart. This information could then be used for further processing and image enhancement. A possible application would be to exploit the high signal-to-noise ratio of MR imaging, especially in soft-tissue regions, to apply denoising methods on the corresponding X-ray images. Considering that the noise level in X-ray Fluoroscopy is directly related to the applied radiation dose, a higher tolerance for noise could lead to reduction of harmful patient radiation exposure. Furthermore, it allows for investigations in the field of super-resolution. 
Most of the mentioned applications would require corresponding images in the same domain. The acquisition of projection images that match the typical projective distortion directly from the MR is possible, as shown by \cite{Napel1991a,Syben2017}. To allow for further down-stream processing, a possibility to transfer the information between the projection images in the distinct domains would be useful.  
Similar methods are already used in radiation therapy planning, where attenuation maps are estimated from pseudo-CT scans that are synthesized from corresponding MR data \cite{Navalpakkam2013, Nie2017, Wolterink2017}. However, all these methods are based on 3D tomographic image data. In contrast, for fluoroscopic procedures this transfer between the domains must be performed based on line integral data, i.e., 2D projection images, and not on reconstructed images. Motivated by its possible applications and inspired by existing methods from radiation therapy and natural image synthesis, we investigate different deep learning-based methods for X-ray projection image synthesis from MR projections. 

\section{Methods}
\label{sec:methods}
Convolutional neural networks have shown great results in natural and medical image synthesis \cite{Nie2017,Gatys2016}. Based on this, three different generator network architectures are used in the underlying work with the goal to generate X-ray projections $\vect{G}$ from input MR projection images $\vect{I}$. Training and evaluation are done using corresponding MR and label X-ray projections $\vect{L}$. All models have been adapted to our specific application. An overview of the investigated network architectures is given in Figure~\ref{fig:architecture}. Furthermore, we examined the impact of two different loss-functions on the generated results. 

\subsection{Model Architecture}
\label{subsec:architecture}
\begin{figure*}
	
	\includegraphics[width=1\textwidth]{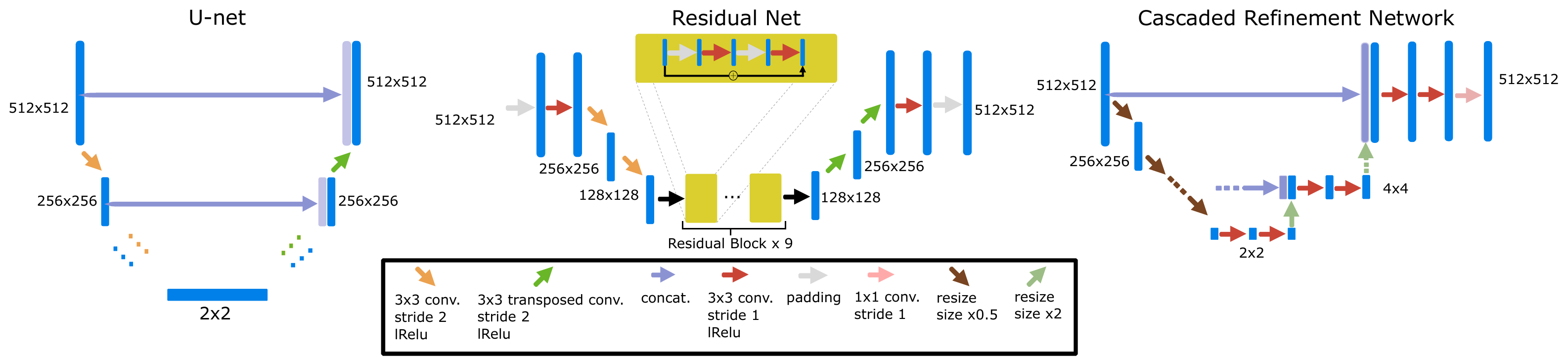}
	\caption{Schematic architectures of the different generator networks.}
	\label{fig:architecture}
\end{figure*}
Convolutional auto-encoders are a popular choice for generator networks in image synthesis. In general, an auto-encoder consists of an encoder and a decoder path. In the encoder path the image's resolution is decreased and the filter dimension is increased. The subsequent decoder path reverts this process to reach the initial resolution and dimension again. Enhancing the encoder-decoder structure with skip-connections between corresponding resolution levels has proven to be beneficial regarding the conservation of spatial information lost during down-sampling. Our first network model is close to the well-known "U-net" introduced by Ronneberger et al. \cite{Ronneberger2015}. Instead of maximum pooling layers we use strided convolution with stride two for up- and down-sampling. In addition to the architecture presented in Figure \ref{fig:architecture}, the first three layers of the synthesis path use dropout with a keep probability  of 50 percent.

The second generator network is a deep residual network (ResNet) \cite{He2015} which was initially proposed for image recognition. The key component of this approach are residual connections that allow for more robust training of deeper networks than before. Besides the original application, this network architecture proved to yield good result in generative tasks. We use the model proposed by \cite{Johnson2016} for style transfer to generate our estimated X-ray projections. Deviating from their proposal, we add nine residual blocks instead of the originally proposed five. 

Finally, a cascaded refinement network (CRN) is used as image generator. This model was recently proposed by Chen et al. \cite{Chen2017} and yielded good results on natural image synthesis from a semantic layout. In contrast to many currently proposed approaches, their model does not use adversarial training but relies on a single feedforward network. The semantic layout as input is replaced by MR projection images in our case. The network consists of multiple refinement modules that work in a multi-scale strategy from coarse to fine as presented in Figure~\ref{fig:architecture}. The full model is built from 8 single refinement modules and the final 1~x~1 convolution layer maps the output to a single channel image. A major difference to the first two network architectures is that Chen et al. relinquished convolutional layers in the down-scaling path and, instead, only use resizing operations. Input information from higher resolution scales is solely incorporated using concatenation. By this, additional model capacity can be used for the subsequent up-scaling path. 
\subsection{Objective Functions}
\label{subsec:objective_func}
The choice of the objective function is a key aspect in every machine learning application. Multiple functions have been used for the task of image-to-image translation and image synthesis in the past. We picked two different loss-functions to compare them in our approach. Since a one-to-one correspondence is given by the matching image pairs, a simple but suitable loss function for image generation tasks is the $\ell_1$-norm \cite{Zhao2017}. Pixel-wise comparison of the generated and label image intensities via the $\ell_1$-loss function can be done by calculating
\begin{equation}
\label{eq:l1norm}
\text{E}_{\ell_1}(\vect{L},\vect{G}) = \sum_{i}^{N} \lvert\vect{L}(i) - \vect{G}(i)\rvert \;,
\end{equation}
where $i$ denotes one image pixel, $i \in N$, and N is the number of all pixel in one image. 

A second loss function that was recently proposed for natural image synthesis without corresponding image pairs is the perceptual-loss \cite{Johnson2016}. The perceptual-loss does not calculate the error between the estimated and real intensity values. Instead, the generated and the label image are fed into a pre-trained image classification network that we will refer to as evaluation network in the following. While the resulting classification scores are not of interest, the raw feature activations between the different input images are compared. The underlying theory is that similarly looking images activate the same units in the image classification network, i.e. the higher the accordance between both feature activations the more similar the generated and label image are. The loss function can be written as
\begin{equation}
\label{eq:perceptual_loss}
\text{E}_\text{p}(\vect{L},\vect{G}) = \sum_{k}^{K} \left( \vect{V}_k(\vect{L}) - \vect{V}_k(\vect{G}) \right) \;,
\end{equation}
where $\vect{V}_k(\vect{L})$ and $\vect{V}_k(\vect{G})$ is the feature activation map of the evaluation network for the label image $\vect{L}$ and the generated image $\vect{G}$ at the current layer $k$, $k \in K$. 
In this approach, the perceptual-loss is computed on the VGG-19 network \cite{Simonyan2015} which was pre-trained on the ImageNet data set \cite{Russakovsky2014}.

All generators are trained with an ADAM optimizer \cite{Kingma2014} and a learning rate of 0.004 for 100 epochs. 
\section{Experiments}
\label{sec:experiments}
Experiments were conducted using data of a realistic MR and X-ray sensitive phantom of the human head. Data was acquired on a 1.5~T Aera MR and a Axiom-Artis C-arm CT scanner (Siemens Healthcare GmbH, Forchheim, Germany). An ultra-short echo time sequence was used for the MRI scans. The reconstructed images' resolution is 320~x~320~x~250 with a spacing of 0.93~x~0.93~x~0.93~mm$^3$. The X-ray scans of the same phantom exhibit a voxel size of 0.48~x~0.48~x~0.48~mm$^3$ and a resolution of 512~x~512~x~399. 
Image registration of the corresponding scans was performed using elastix. The input (MR) and label (CT) images were generated by forward projecting the registered stack from various angulations using the CONRAD framework \cite{Maier2013}. In this manner, 3200 different projection image pairs of both modalities were created and randomly divided into 3000 training and 200 testing images.

The evaluation of the output can be done by calculating the deviation of the generated X-ray $\vect{G}$ from the real X-ray images $\vect{L}$. The mean squared error (MSE) can be used to this end. It is \mbox{computed} as 
\begin{equation}
\label{eq:mse}
\text{MSE}(\vect{L}, \vect{G}) = \frac{1}{N}\sum_{i}^{N} \big\| \vect{L}(i) - \vect{G}(i) \big\|^2_2 \;.
\end{equation}
Yet, not only the absolute difference of estimated values is of interest in projection image synthesis. The generated projection images must also correspond to each other from a visual point of view, which cannot be determined entirely be pixel-wise comparison of the image pairs. To this end, the structural similarity (SSIM) index \cite{Wang2004}, a perception-based metric, is computed. Assuming two patches $\vect{g}$ and $\vect{l}$ of the generated and label image. The SSIM is then computed as
\begin{equation}
\text{SSIM}(\vect{g},\vect{l}) = \frac{(2\mu_{\vect{g}}\mu_{\vect{l}} + c_1)(2\sigma_{\vect{g}\vect{l}}+c_2)}{(\mu^2_{\vect{g}} + \mu^2_{\vect{l}} + c_1)({\sigma^2_{\vect{g}}} + {\sigma^2_{\vect{l}}} + c_2)}\;, 
\label{eq:ssim}
\end{equation}
where $\mu$ is the mean, $\sigma^2$ the variance, and $\sigma$ the covariance. To avoid instabilities, the constants $c_1$ and $c_2$ are introduced that are defined as $c_i = (K_i\mathcal{L})^2$, $i\in{\{1,2\}}$, with $\mathcal{L}$ being the dynamic range of the intensity values and $K_1 = 0.01$ and $K_2 = 0.03$. Computing Equation \ref{eq:ssim} for all pairs of patches $\vect{g}$ and $\vect{l}$ yields the final SSIM measure for the whole image.  

The third evaluation metric that is computed is the peak signal-to-noise ratio (PSNR). The PSNR measures the ratio between the highest intensity value and the occuring noise and is often applied to measure image quality, especially regarding reconstruction and compression loss. It is computed by
\begin{equation}
\text{PSNR}(\vect{L}, \vect{G}) = 20\text{log}_{10}\frac{\text{max}(\vect{G})}{\text{MSE}(\vect{L}, \vect{G})} \; .
\label{eq:psnr}	
\end{equation}
In the subsequent chapter results for all metrics will be presented. To present comparable absolute numbers, all images were scaled from -1 to 1 prior to the error metric calculations.
\section{Results and Discussion}
\label{sec:results}
\begin{table}
	\includegraphics[width=0.45\textwidth]{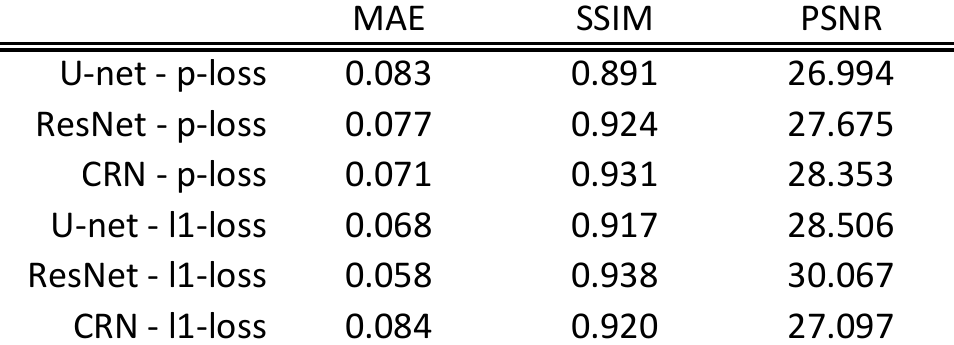}
	\caption{Quantitative results of the different network architectures and loss functions}
	\label{tab:results_projection_synthesis}
\end{table}
\begin{figure*}
	\centering
	\subfloat[Input: MR proj.]{\includegraphics[width=0.19\textwidth]{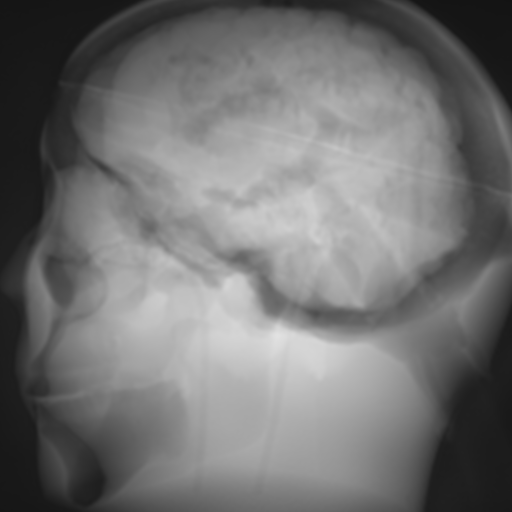}%
		\label{fig:input_3000}}
	\hspace{1pt}
	\subfloat[Output: U-net p-loss.]{\includegraphics[width=0.19\textwidth]{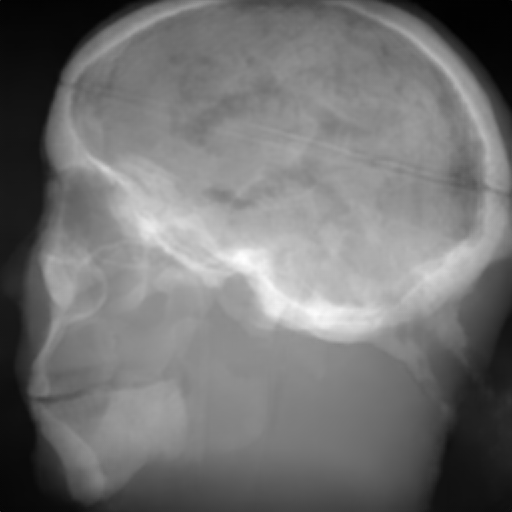}%
		\label{fig:unet_FM_3000}}
	\hspace{1pt}
	\subfloat[Output: ResNet p-loss.]{\includegraphics[width=0.19\textwidth]{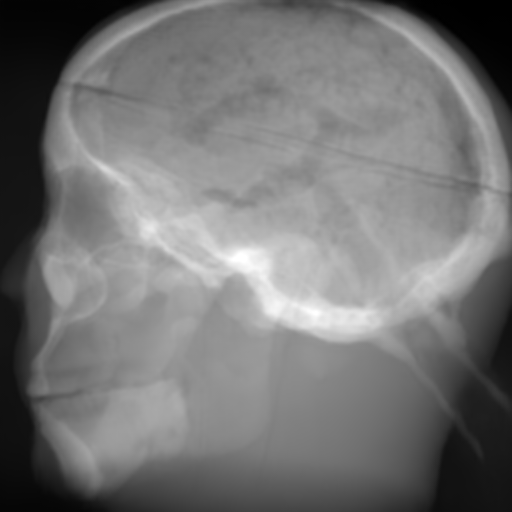}
		\label{fig:resnet_FM_3000}}
	\hspace{1pt}
	\subfloat[Output: CRN p-loss.]{\includegraphics[width=0.19\textwidth]{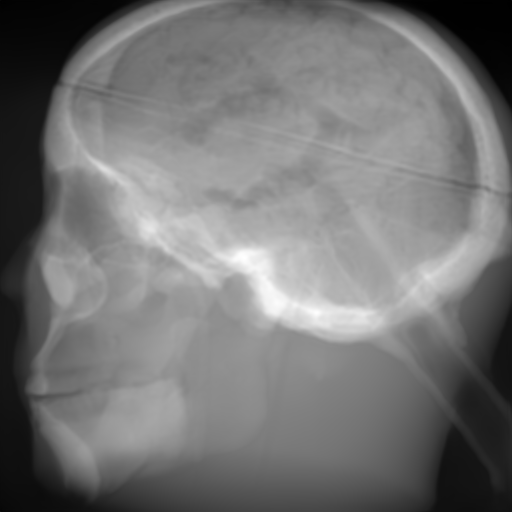}%
		\label{fig:recgen_FM_3000}}
	\hspace{1pt}
	\subfloat[Reference: X-ray proj.]{\includegraphics[width=0.19\textwidth]{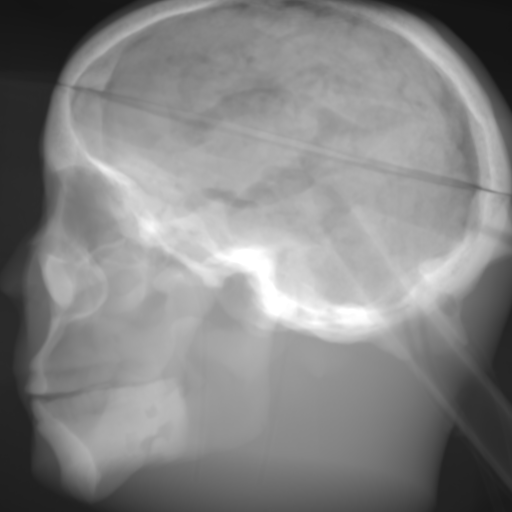}%
		\label{fig:label_3000}}
	
	\subfloat[Input: MR proj.]{\includegraphics[width=0.19\textwidth]{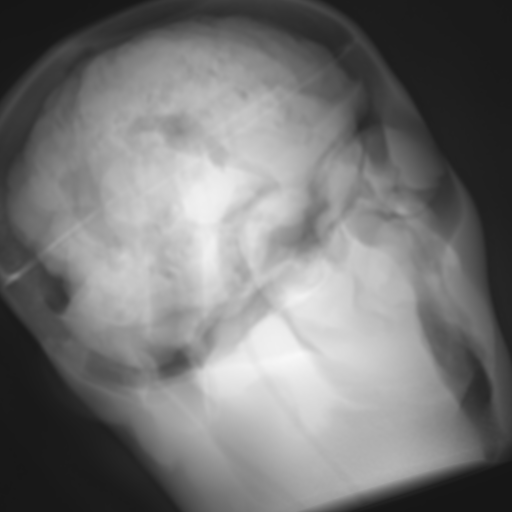}%
		\label{fig:input_3076}}
	\hspace{1pt}
	\subfloat[Output: U-net $\ell_1$-loss.]{\includegraphics[width=0.19\textwidth]{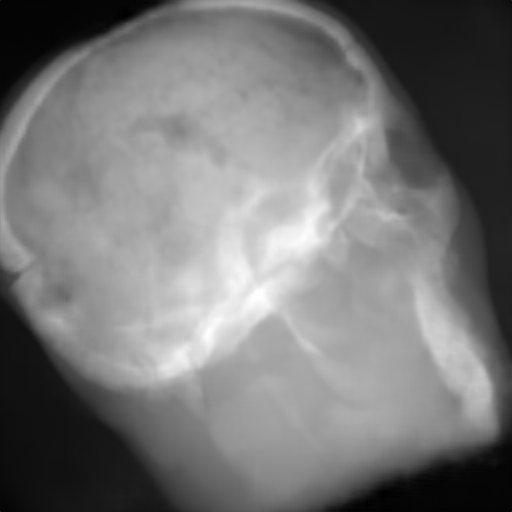}%
		\label{fig:unet_l1_3076}}
	\hspace{1pt}
	\subfloat[Output: ResNet $\ell_1$-loss.]{\includegraphics[width=0.19\textwidth]{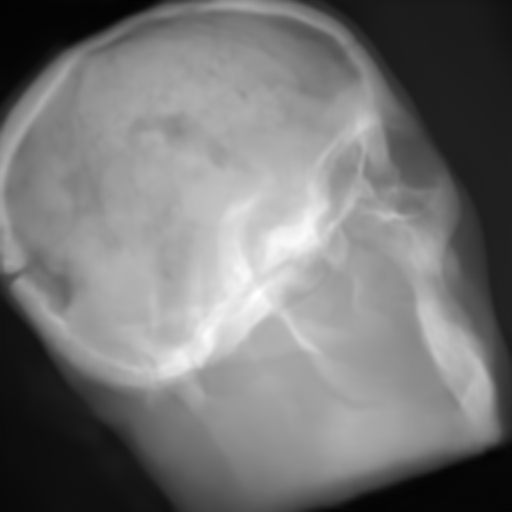}%
		\label{fig:resnet_l1_3076}}
	\hspace{1pt}
	\subfloat[Output: CRN $\ell_1$-loss.]{\includegraphics[width=0.19\textwidth]{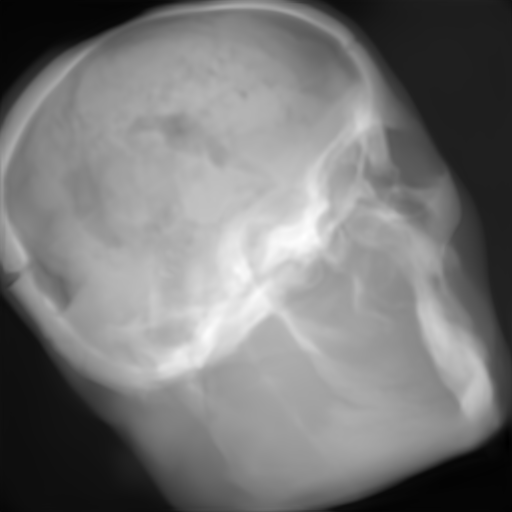}%
		\label{fig:recgen_l1_3076}}
	\hspace{1pt}
	\subfloat[Reference: X-ray proj.]{\includegraphics[width=0.19\textwidth]{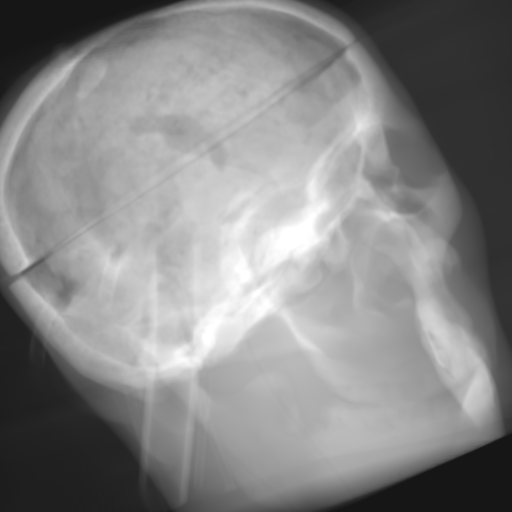}%
		\label{fig:label_3076}}
	\caption[Projection synthesis results]{Results of the projection synthesis. Top row: Results generated with the perceptual-loss function. \mbox{Bottom row:} Results generated with the $\ell_1$-loss function.}
	\vspace{-0.3cm}	
	\label{fig:proj_synthesis_results} 
\end{figure*}
The quantitative and qualitative results of the proposed experiments are presented in Table \ref{tab:results_projection_synthesis} and Figure \ref{fig:proj_synthesis_results}. By examining these it can be observed that the differences in the calculated MSE of all network architectures and incorporated loss functions are only small. The best results in terms of pixel-wise deviation could be achieved with the ResNet architecture combined with the $\ell_1$-loss function. This network achieves a deviation from the reference of only 0.058, i.e., 2.4 percent. Also the results of the U-net and CRN networks are still good with deviations of 2.6 and 2.9 percent. Similarly small variation can be observed in the structured similarity measure. 
The ResNet and CRN exhibit approximately equal quality with SSIM measures of 0.938 and 0.920 for the $\ell_1$-loss and 0.924 and 0.931 for the perceptual-loss, respectively. The results generated with the U-net are slightly worse. The highest peak signal-to-noise ratio is achieved by the ResNet ($\ell_1$-loss), followed by the U-net ($\ell_1$-loss) and CRN (p-loss). It is noteworthy that the ResNet and U-net both achieve the highest results in all error metrics using the  $\ell_1$-loss while the opposite is the case for the CRN which works best with the perceptual-loss function.

Overall, the perceptual-loss achieves competitive and in some cases even better results than the $\ell_1$-loss when comparing the pixel-wise error metrics. For example, the cascaded refinement network's MSE is 0.013 smaller for the perceptual- than for the $\ell_1$-loss. This might be suspicious at first sight, considering that the $\ell_1$-loss purely optimizes for this pixel-wise error in the training process while the perceptual-loss compares the raw feature activations of the evaluation network. Contrarily, this behavior cannot be observed for the U-net and ResNet. The results produced with the $\ell_1$-loss achieve higher values for all error measures for these networks. An explanation for this obervation is that the intensity values of the input image still cause an impact on the respective layers output in the evaluation network when computing the perceptual-loss. Consequently, these differences also transition to the computed loss value for all feature layers. Even though the perceptual-loss incorporates the raw intensity values, it is not guaranteed that the scaling of these is conserved in this process. By this, the relative changes can be similar, whereas the absolute range of values changes and, correspondingly, also the pixel-wise error metrics.  

Another observation is that the perceptual-loss is able to conserve high-frequency details in the image. The fine line in the projection images that forms a circle around the cranium is visible in the input \mbox{(Figures \ref{fig:input_3000}~\&~\ref{fig:input_3076})}, as well as in the label images (Figures~\ref{fig:label_3000}~\&~\ref{fig:label_3076}), and also in the images generated with the perceptual-loss function (Figures~\ref{fig:unet_FM_3000},~\ref{fig:resnet_FM_3000},~and~\ref{fig:recgen_FM_3000}). In contrast, all generators "loose" this line when the $\ell1$-loss is applied (Figures \ref{fig:unet_l1_3076},~\ref{fig:resnet_l1_3076},~and~\ref{fig:recgen_l1_3076}). This effect is also qualitatively observable in other parts of the images. Despite achieving equal or better results regarding the error metrics, the generally less sharp look of the results generated with the $\ell1$-loss function is apparent. This behavior is in accordance with previous observations that concluded that an perceptual-loss leads to sharper images than a comparable $\ell1$-loss \cite{Dosovitskiy2016}. Considering the common applications of X-ray Fluoroscopy, e.g., interventional guidance for stents and similar devices, high spatial resolution is a key requirement. Utilizing a loss function that is able to preserve high-frequency details in the images is desirable to this end. The perceptual-loss appears to be suited for this task as presented in our evaluation.

\section{Conclusion}
\label{sec:subhead}
We showed the feasibility of image-to-image translation from MR projection images to corresponding X-ray projections. Three generator networks and two different loss functions were implemented and evaluated to this end. All examined network architectures achieved good results on the proposed task. When comparing the generated projection images of all networks it became apparent that the loss function has a greater impact on the images' quality than the actual architectures of the network. The perceptual-loss proved to be able to conserve even small high-frequency details in the course of the image-to-image transfer. Because high-spatial resolution is desired in most fluoroscopic procedures, we recommend using this perceptual-loss function for the underlying task. The best quantitative and qualitative results with this loss function could be achieved by a cascaded refinement model in this work. The high-quality of the generated projection images unveils large potential regarding the applicability to multi-modal denoising, super-resolution, and more. As a next step, we plan to transfer this approach to real patient data. Additionally, the effect of combining multiple different MR acquisition protocols and weighting schemes will be investigated. 

\bibliographystyle{IEEEbib}
\bibliography{lib/library}

\end{document}